\documentclass{llncs}
\usepackage{amssymb,amsmath}
\usepackage{ifxetex,ifluatex}
\ifxetex
  \usepackage{fontspec,xltxtra,xunicode}
  \defaultfontfeatures{Mapping=tex-text,Scale=MatchLowercase}
\else
  \ifluatex
    \usepackage{fontspec}
    \defaultfontfeatures{Mapping=tex-text,Scale=MatchLowercase}
  \else
    \usepackage[utf8]{inputenc}
  \fi
\fi
\makeatletter\AtBeginDocument{%
  \renewcommand{\@listi}
    {\setlength{\labelwidth}{4em}}
}\makeatother
\usepackage{enumerate}
\usepackage{ctable}
\usepackage{float} 
\ifxetex
  \usepackage[setpagesize=false, 
              unicode=false, 
              xetex]{hyperref}
\else
  \usepackage[unicode=true]{hyperref}
\fi
\hypersetup{breaklinks=true, pdfborder={0 0 0}}
\title{Learning Sequence Neighbourhood Metrics}
\author{Justin Bayer, \email{bayer.justin@googlemail.com}
\thanks{Contacting author.}
\inst{1}
\and Christian Osendorfer, \email{osendorf@in.tum.de}
\inst{1}
\and Patrick van der Smagt, \email{smagt@dlr.de}
\inst{2}
}

\institute{Chair for Robotics and Embedded Systems, Insitut f\"ur Informatik, Technische Universit\"at M\"unchen \and 
Institute of Robotics and Mechatronics, DLR  German Aerospace Center
}

\begin{document}
\maketitle

\begin{abstract}
Recurrent neural networks (RNNs) in
combination with a pooling operator and the neighbourhood components analysis
(NCA) objective function are able to detect the characterizing dynamics of
sequences and embed them into a fixed-length vector space of arbitrary
dimensionality.
Subsequently, the resulting features are meaningful and can be used for
visualization or nearest neighbour classification in linear time.
This kind of metric learning for sequential data enables the use of
algorithms tailored towards fixed length vector spaces such as $\mathbb{R}^n$.
\end{abstract}

\section{1. Introduction}

Sequential data is found in many domains including medical applications,
robot control, neuroscience, financial information or text processing.
This data is fundamentally different from static data vectors.

When considering a single sequence over $T$ time steps
$\mathbf{x} = (x_1, x_2, \dots, x_T) \in X^*$ with
$X \subset \mathbb{R}^n$, the order of the individual elements $x_i$ is
relevant for the interpretation\footnote{We let $X^*$ denote the set of
  all sequences over the space $X$.}. Conversely, in the case of static
data $\mathbf{x'}\in \mathbb{R}^n$, an ordering on the $n$ components is
not even defined. Indeed, the key element of structured data is that the
context (i.e., the immediate predecessors and successors) contains
essential information to make learning on the data possible.

 \cite{nca} note that metrics and features are actually closely related:
by measuring pairwise distances between the data points
$\mathbf{x}' \in \mathbb{R}^n$, the data can be embedded into a metric
space. They learn a Mahalanobis distance by mapping the high-dimensional
data set $X$ to a metric space $Z$ in which $k$-nearest neighbour
classification performance is maximized. The resulting objective
function is differentiable with respect to the embedding.

Similar to \cite{nonlinear-nca}, we use a different model than a linear map
for learning the
embedding function. Our choice, recurrent neural networks (RNNs), are
rich models for sequence learning. They have been successfully used for
handwriting recognition \cite{write-rnn}, audio processing
\cite{speech-rnn}, and text modelling \cite{text-rnn}.

\subsection{Related Work}

Only a few principled approaches exist for extracting fixed length
features from sequential data. If we were given some kind of distance
measure, a classic technique such as multi-dimensional scaling could be
used. This is however rarely the case. A commonly used practice is based
on a set of fixed basis functions (e.g.~Fourier or wavelet basis). While
it has strong mathematical guarantees, it is sometimes too inflexible:
in order to work with arbitrarily long sequences, a sliding time window
has to be employed, limiting the capability to model context.
Furthermore, the fixed set of basis functions implies that the problem
of identifying usefull factors of variations remains unresolved in
general. Fisher kernels \cite{Jaakkola98exploitinggenerative}, a
combination of probabilistic generative models with kernel methods,
provide another commonly vectorial representation of sequences. The
basic idea is that two similar objects induce similar gradients of the
likelihood for the parameters of the model. Thus, the features for a
sequence are the elements of the gradient of the log-likelihood of this
sequence with respect to the model paramters. This choice can presumably
be very bad: if the distribtution represented by the trained model
closely resembles the data distribution the gradients for all sequences
in the data set will be nearly zero. A recent paper
\cite{learn-fisher-kernels} alleviates this problem by exploiting label
information and employing ideas from metric learninig. Obviously, this
only works if class information is available.

A fully unsupervised approach is to use the parameters estimated by a
system identification method (e.g., a linear dynamical system) as
features. Recent work includes \cite{complex-lds}, in which a complex
numbers based system successfully clusters motion capture data.

The last two approaches clearly suffer from the fact that the number of
features is directly connected with the complexity of the model. In
particular it is not given that the important factors of variation are
captured by these methods.

In principle, any sequenctial clustering technique can be used as a feature
extractor by treating the scores (e.g. the posterior likelihood in case of a
generative model or the distances to a node of a self-organizing map)
as features.

\section{2. Recurrent Neural Networks}

Recurrent neural networks are an extension of feedforward networks to
deterministic state space models. The inputs to an RNN are given as a
sequence $(x_1, x_2, \dots, x_T)$. Subsequently, a sequence of hidden
states $(h_1, h_2, \dots, h_T)$ and a sequence of outputs
$(o_1, o_2, \dots, o_T)$ is calculated via the following equations:
\begin{eqnarray}
h_t & = & \sigma(W_{xh} x_t + W_{hh} h_{t-1} + b_h) \\
 o_t & = & W_{ho}h_t + b_o
\end{eqnarray}

where $t=1, 2, \dots, T$ and $\sigma$ is a suitable transfer function,
typically the tangent hyperbolic, applied element-wise. $W_{\Diamond}$
are weight matrices, $b_{\Diamond}$ bias vectors and $x_i, h_i$ and
$o_i$ real-valued vectors. For the calculation of $h_1$ a special
initial hidden state $h_0$ has to be used which can be optimized during
learning as well.

RNNs have a lot of expressive power since their states are distributed
and nonlinear dynamics can be modelled. The calculation of their
gradients is astonishingly easy via Backpropagation Through Time (BPTT)
\cite{bptt} or Real-Time Recurrent Learning \cite{rtrl}. However, first
order gradient methods completely fail to capture relations that are
more than as little as ten time steps apart of each other. This problem
is called the \emph{vanishing gradient} and has been studied by
\cite{vanishing-gradient-sepp} and \cite{vanishing-gradient-bengio}. The
previous state of the art method to overcome this has been the Long
Short-Term Memory (LSTM) \cite{lstm} up until recently, when
\cite{hf-rnn} introduced a second-order optimization method for RNNs, a
Hessian free optimizer (HF-RNN), which is able to cope with
aforementioned long term dependencies even better. In this work, we
stick to LSTM since the HF-RNN is tailored towards convex loss
functions---neighbourhood component analysis (NCA), the objective
function of choice in this paper, is however not convex.

Another neural model for nonlinear dynamical systems is the echo state
network approach introduced in \cite{echo-state}. The drawback of this
method is that the dynamics that are to be modelled have to be already
present in the network's random initialization.

\subsection{Recurrent Networks are Differentiable Sequence
Approximators}

One consequence of the differentiability of RNNs is that we can optimize
their parameters with respect to an objective function.\footnote{The
  authors recommend to use automatic or symbolic differentiation. In
  this work, Theano \cite{theano} was used.} Stochastic gradient descent
or higher order techniques are the techniques of choice to fit the
weights.

Similar to \cite{collobert:2011b} we reduce output sequences to a single
vector with a \emph{pooling operation}. A pooling operation is a
function $p: X^* \rightarrow X$ that reduces an undefined amout of
inputs to a single output of the same set, e.g.~taking the sum or
picking the maximum. Similar to convolutional neural networks, we can
use this technique to reduce a sequence to a point. If our pooling
operation is differentiable as well, we can use it as a gateway to
arbitrary objective functions that are defined on real vectors. Given a
network $f$ parametrized by $W$, a data set
$\mathcal{D}=\lbrace x_i \rbrace$, a pooling operation $p$ and an
objective function $\mathcal{O}$ we proceed as follows:

\begin{enumerate}[1.]
\item
  Process input sequences
  $\mathbf{x}_i = (x_{i1}, \dots, x_{iT}), x_{it} \in \mathbb{R}^n$ to
  produce output sequences
  $f(\mathbf{x}_i; W) = \mathbf{o}_i = (o_{i1}, \dots, o_{iT}), o_{it} \in \mathbb{R}^m$,
\item
  Use a pooling operation $p$ to reduce the output sequences to a point
  via $p(o_{i1}, \dots, o_{iT}) = e_i$,
\item
  Calculate the objective function $\mathcal{O}(\lbrace e_i \rbrace)$.
\end{enumerate}
Since the whole calculation is differentiable, we can evaluate the
derivative of the objective function with respect to the parameters of
the RNN via
\begin{eqnarray}
\frac{\mathcal{\partial{O}}}{\partial{p}} \frac{\partial{p}}{\partial{f}} \frac{\partial{f}}{\partial{W}}. \label{eqn-whole-deriv}
\end{eqnarray}

Subsequently, we can use the gradients to find embeddings
$\lbrace e_i \rbrace$ of our data which optimize the objective function.
We apply this insight to combine RNNs with neighbourhood components
analysis (NCA), which we will review in a later section.

\subsection{Long Short-Term Memory}

LSTM cells are special stateful transfer functions for RNNs which enable
the memorization of events hundreds of time steps apart. We review them
shortly because the usage of LSTM cells plays a crucial role in problems
where long term dependencies are an important characteristic of the data
at hand, necessary to make usable predictions.

The power of LSTM cells is mostly
attributed to a special building block, the so called
\emph{gating units}. We define $\phi(c, v) = v\sigma(c)$ with $\sigma$
being the sigmoid function $\frac{1}{1 + e^{-x}}$ ranging from $0$ to
$1$.

Another central concept are the \emph{states} $(s_1, s_2, \dots, s_T)$
of the cell. These can be altered by the inputs via the \emph{input},
\emph{forget} and \emph{output gate}. To keep the notation uncluttered,
we concatenate the four different inputs $a_t^{(\cdot)}$ to the cell
into a single vector. As indicated by the superscript, each of the
$a_t^{(\cdot)}$ represents an input to one of the gates $i$, $f$ and
$o$. The superscript $x$ represents the input to the cell itself.
\begin{eqnarray*}[a^{(x)}_t \: a^{(i)}_t \:  a^{(f)}_t \:  a^{(g)}_t] & = & W_{ha}h_{t-1} + W_{xa}x_t + b_a \label{lstm-global}\\
s_t & = & \underbrace{\phi(a^{(i)}_t, a^{(x)}_t)}_{\text{input gate}}
                 + \underbrace{\phi(a^{(f)}, s_{t-1})}_{\text{forget gate}} \label{lstm-state} \\
h_t & = & \sigma(\underbrace{\phi(a^{(o)}_t, s_t)}_{\text{output gate}}) \label{lstm-hidden} \\
o_t & = & W_{ho} h_t + b_h \notag
\end{eqnarray*}

Since all the operations are differentiable, gradient-based can be employed.

\section{3. Sequential Neighbourhood Components Analysis}

The central assumption of neighbourhood components analysis
\cite{nca,nonlinear-nca} is that items of the same class lie near each
other on a lower-dimensional manifold. To exploit this, we want to learn
a function $f: X^* \rightarrow Z$ from the sequence space $X$ to a
metric space $Z$ that reflects this. Recall that in our case, the
embedding function is $e(\mathbf{x}; W) = p(f(\mathbf{x}; W))f$. Given a
set of sequences with an associated class label
$\mathcal{D} = \lbrace x_i, c_i \rbrace$ mapped to a set of embeddings
$\mathcal{E} = \lbrace e_i \rbrace$, we define the probability that a
point $a$ selects another point $b$ as its neighbour based on Euclidean
pairwise distances as
\begin{eqnarray*}
    p_{ab} = \frac{\exp(-||e_a - e_b||^2_2)}{\sum_{z \neq a}\exp(-||e_a - e_z||^2_2)},
\end{eqnarray*}

while the probability that a point selects itself as a neighbour is set
to zero: $p_{aa} = 0$. The probability that a point $i$ is assigned to
class $k$ depends on the classes of the points in its neighbourhood
$p(c_i = k) = \sum_j p_{ij} \mathbb{I}(c_j = k)$, where $\mathbb{I}$ is
the indicator function.function. The overall objective function is then
the expected number of correctly classified points
\begin{eqnarray*}
    \mathcal{O} = \sum_i \sum_j p_{ij} \mathbb{I}(c_i = c_j).
\end{eqnarray*}

Although NCA has a computational complexity that is quadratic in the
number of samples in the training set for training, using batches
containing roughly 1000 samples made this negligible. We did not observe
any decrease of test performance.

\subsection{Classifying Sequences}

We first train an RNN on our data set with the NCA objective function.
Afterwards, all training sequences are propagated through the network
and the pooling operator to obtain embeddings
$\mathcal{E} = \lbrace e_i \rbrace$ for each of them. We then build a
nearest neighbour classifier for which we use all embeddings of the
training set. A new sequence $(x_1, x_2, \dots, x_T)$ is classified by
first forward propagating it through the RNN and obtaining an embedding.
We then find the $k$-nearest neighbours and obtain the class by a
majority vote.

Note that this method has two appealing characteristics from a
computational perspective: first, finding a descriptor for a new
sequence has a complexity in the order of the length of that sequence.
Furthermore, the memory requirements for that descriptor are invariant
of the length of the sequence and can thus be tailored towards memory
requirements. Indeed, millions of such descriptors can easily be held in
main memory to allow fast similarity search.

\section{4. Experiments}

To show that our algorithm works as a classifier we present results on
several data sets from the UCR Time Series archive
\cite{ucr-time-series-hp}. Due to space limitations, we refer the reader
to the corresponding web page for detailed descriptions of each data
set. The data sets from UCR are restricted in the sense that all are
univariate and of equal length. Since our method is well suited to high
dimensional sequences, we we proceed to the well known TIDIGITS
benchmark afterwards.

\subsection{UCR Time Series Data}

The hyper parameters for each experiment were determined by random
search. We did 200 experiments for each data set, reporting the test
error for those parameters which performed best on the training set. The
hyper parameters were the number of hiddens, the used transfer function
(sigmoid, tangent hyperbolicus rectified linear units or lstm cells),
the optimization algorithm (either RPROP or LBFGS), the pooling operator
(either sum, max or mean), whether to center and whiten each sequence or
the whole data set and the size of the batch to perform gradient
calculations on.

\small
\begin{center}
\begin{tabular*}{\textwidth}[hc]{@{\extracolsep{\fill}}lcccc}
\hline
Data set & Train & Test & our 1NN & DWT 1NN \\
\hline
Wafers & 0.984 & 0.987 & 0.987 & 0.995
\\
Two Patterns & 0.992 & 0.996 & 0.99725 & 0.9985
\\
Swedish Leaf & 0.797 & 0.772 & 0.848 & 0.843
\\
OSU Leaf & 0.684 & 0.457 & 0.579 & 0.616
\\
Face (all) & 0.938 & 0.833 & 0.647 & 0.808
\\
Synthetic Control & 0.999 & 0.962 & 0.96 & 0.983
\\
ECG & 0.999 & 0.846 & 0.88 & 0.88
\\
Yoga & 0.684 & 0.73 & 0.699 & 0.845 \\
\end{tabular*}
\end{center}

The training and test performances stated are the average probabilities
that a point is correctly classified by the stochastic classifier used
in the formulation of NCA. We also report the error for 1-nearest
neighbour classification on the test set as 1NN with the training set as
a data base to perform nearest neighbour queries on. 1NN-DWT corresponds to the
best DWT classification results on the UCR page. If a certrain data set from the
UCR repository is not listed, performance was not satisfactory. We attribute
this to small training set sizes in comparison with the number of classes with
which our method seems to struggle. This is not at all
surprising, as the number of parameters is sometimes exceeded by the number of
training samples.

\subsection{TIDIGITS Data}

TIDIGITS is a data set consisting of spoken digits by adult and child
speakers. We restricted ourselves to the adult speakers. The audio was
preprocessed with mel-frequency cepstrum coefficient analysis to yield a
13-dimensional vector at each time step.

\begin{figure}[t]
\begin{center}
\includegraphics[width=6cm]{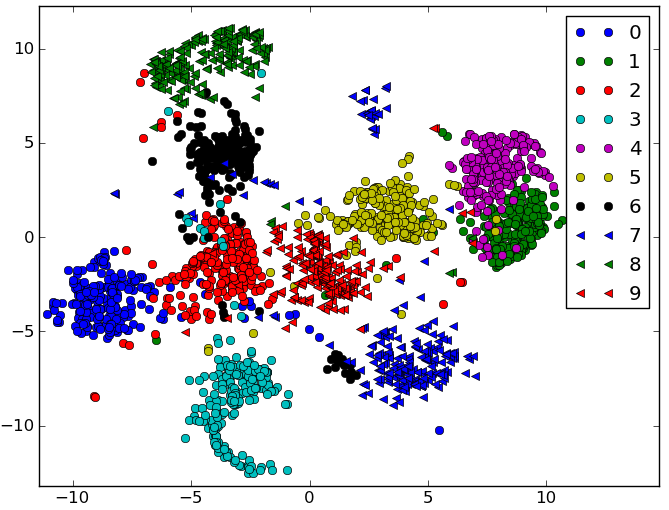}
\end{center}
\caption{The output of our method after applying tSNE to the found embeddings.
The data is arranged into mostly distinct clusters. Interestingly, the NCA
objective also makes it possible for points of the same class to arrange in
several clusters. This is not the case for objectives that try to separate the
data with a functional form such as a hyperplane.}
\label{res-tidigits}
\end{figure}

During training we went along with the official split into a set of 2240
training and 2260 testing samples. 240 samples from the training set
were used for validation. We trained the networks until convergence and
report the test error with the parameters achieving the best validation
error. We used 40 LSTM \cite{lstm} units to get 30 dimensional
embeddings. For comparison, we also trained LSTM-RNNs of similar size
with the cross entropy error function for comparison. Since both methods
yield discriminative models, we can report the the average probability
that a point from the testing set is correctly classified, which was
97.9\% for NCA and 92.6\% for cross entropy. For a visualization of the
found embeddings, see figure \ref{res-tidigits}.

\section{5. Conclusion}

We presented a solution to an important problem---by combining two well
established methods we introduced a method to embed sequential data into
a semantically meaningful feature space: it leads to interpretable
features naturally and can be used out of the box as a visualization
method and data exploration tool.

The techniques presented here are usable with any RNN structure---we
believe that the usage of echo state networks \cite{echo-state} or
multiplicative RNNs \cite{text-rnn} to NCA might yield even better
results.

We also want to stress the applicability of our method to big data:
while classifcation has the downside of quadratic complexity, the
resulting embeddings are extremly well compressed representations of the
data. Also, finding a new representation for an unseen sequence is a
single forward pass of an RNN, which is extremly efficient.

\bibliographystyle{plain}
\bibliography{lib}

\end{document}